\documentclass[letterpaper, 10 pt, conference]{ieeeconf}  % Comment this line out if you need a4paper

\IEEEoverridecommandlockouts                              % This command is only needed if 
\title{\LARGE \bf
Accelerated Feature Detectors for Visual SLAM: A Comparative Study of FPGA vs GPU}

\author{Ruiqi Ye and Mikel Luj{\'a}n   % stops a space
\thanks{Ruiqi Ye and Mikel Luj{\'a}n are with Department of Computer Science, 
        University of Manchester, Manchester M13 9PL, UK
        {\tt\small first.last@manchester.ac.uk}}%
}
\usepackage{soul}
\usepackage{amsmath,amssymb,amsfonts}
\usepackage{algorithmic}
\usepackage{graphicx}
\usepackage{textcomp}
\usepackage{xcolor}
\usepackage[ruled,vlined,linesnumbered]{algorithm2e}
\usepackage{makecell}
\usepackage{multirow}
\usepackage[flushleft]{threeparttable} % http://ctan.org/pkg/threeparttable
\usepackage[
backend=biber,
style=numeric,
citestyle=ieee,
sorting=none
]{biblatex}
\usepackage{tikz}
\usepackage{pifont}
\def\checkmark{\tikz\fill[scale=0.4](0,.35) -- (.25,0) -- (1,.7) -- (.25,.15) -- cycle;}
\newcommand{\xmark}{\ding{55}}
\begin{document}

\maketitle
\thispagestyle{empty}
\pagestyle{empty}

%%%%%%%%%%%%%%%%%%%%%%%%%%%%%%%%%%%%%%%%%%%%%%%%%%%%%%%%%%%%%%%%%%%%%%%%%%%%%%%%
\begin{abstract}
Feature detection is a common yet time-consuming module in Simultaneous Localization and Mapping (SLAM) implementations, which are increasingly deployed on power-constrained platforms, such as drones. Graphics Processing Units (GPUs) have been a popular accelerator for computer vision in general, and feature detection and SLAM in particular. 

On the other hand, System-on-Chips (SoCs) with integrated Field Programmable Gate Array (FPGA) are also widely available. This paper presents the first study of hardware-accelerated feature detectors considering a Visual SLAM (V-SLAM) pipeline. We offer new insights by comparing the best GPU-accelerated FAST, Harris, and SuperPoint implementations against the FPGA-accelerated counterparts on modern SoCs (Nvidia Jetson Orin and AMD Versal).  

The evaluation shows that when using a non-learning-based feature detector such as FAST and Harris, their GPU implementations, and the GPU-accelerated V-SLAM can achieve better run-time performance and energy efficiency than the FAST and Harris FPGA implementations as well as the FPGA-accelerated V-SLAM. However, when considering a learning-based detector such as SuperPoint, its FPGA implementation can achieve better run-time performance and energy efficiency (up to 3.1$\times$ and 1.4$\times$ improvements, respectively) than the GPU implementation. The FPGA-accelerated V-SLAM can also achieve comparable run-time performance compared to the GPU-accelerated V-SLAM, with better FPS in 2 out of 5 dataset sequences. When considering the accuracy, the results show that the GPU-accelerated V-SLAM is more accurate than the FPGA-accelerated V-SLAM in general. Last but not least, the use of hardware acceleration for feature detection could further improve the performance of the V-SLAM pipeline by having the global bundle adjustment module invoked less frequently without sacrificing accuracy.

\end{abstract}
%\keywords
%Visual-Inertial SLAM, Performance Evaluation and Benchmarking, Software-Hardware Integration for Robot Systems, Embedded Systems for Robotic and Automation
%\endkeywords

%%%%%%%%%%%%%%%%%%%%%%%%%%%%%%%%%%%%%%%%%%%%%%%%%%%%%%%%%%%%%%%%%%%%%%%%%%%%%%%%
\section{Introduction}

Image feature detection has been an important research direction in computer vision and robotics and plays a foundational part in other more complex algorithms, such as image classification, object detection, Visual Odometry (VO), and Simultaneous Localization And Mapping (SLAM). Such tasks sometimes need to be deployed on edge platforms, such as autonomous drones and robots. Edge platforms often operate using batteries and, thus, are constrained by their energy efficiency. However, edge platforms that use only System-on-Chips (SoCs) with embedded microcontrollers (e.g., Arm, RISC-V) tend not to be able to meet the demanding requirements of these complex tasks \cite{gkeka_reconfigurable_2022}.

On the other hand, high-end Graphics Processing Units (GPUs) are accelerators widely used by computer vision and robotics researchers to enable real-time performance \cite{10.1145/3054739}. Embedded GPUs integrated into modern SoCs, such as the Nvidia Jetson family of products \cite{noauthor_nvidia_nodate}, have enabled more energy-efficient robot systems.

Energy-efficient SoCs with embedded GPUs are not the only edge platforms that can accelerate feature detection. SoCs with integrated Field Programmable Gate Array (FPGA), are also widely available. SoCs with integrated FPGAs enable bespoke hardware acceleration for specific algorithms without having to transfer data over PCIe/CXL. This kind of platform has not been studied to the same extent as GPUs for the hardware acceleration of V-SLAM.

\begin{figure}[bt]
    \centering
    \includegraphics[width=1.0\linewidth]{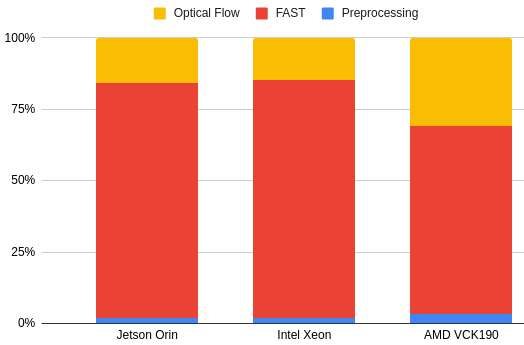}
    \caption{Profiling of the ICE-BA localization thread on an Nvidia Jetson Orin SoC, an Intel Xeon workstation, and an AMD VCK190 SoC using only the processors and Machine Hall sequences. Best viewed in color.}
    \label{fig:ice_ba_frontend_breakdown}
\end{figure}

Despite many advances in the literature, feature detection remains computationally intensive as these algorithms tend to iterate over every pixel in the image, extracting feature points. For example, the popular ICE-BA \cite{liu_ice-ba_2018} uses the FAST \cite{10.1007/11744023_34} feature detector for the front-end of its Visual SLAM (V-SLAM) pipeline. Figure \ref{fig:ice_ba_frontend_breakdown} illustrates the profiling results of the ICE-BA localization thread using the Machine Hall sequences of the EuRoC dataset, on the processors of an Nvidia Jetson Orin, an AMD VCK190, and a workstation with an Intel Xeon chip. The run-time is broken down into three modules: pre-processing, FAST feature detector, and sparse optical flow. On both platforms, the run-time of the FAST detector dominates the run-time, at least 66\% of the execution time. For the other EuRoC MH sequences, the run-time breakdown is similar, and thus, further breakdowns are omitted. For example, with Jetson Orin, the FAST detector always takes up around 80\%-85\% of the run-time. 

Thus, designing hardware accelerators for feature detectors using FPGAs (\cite{aguilar-gonzalez_robust_2018, kalms_fpga_2016, chen_bucket-stream_2020, kalms_full-hd_2018, fischer_rotation_2011, 9725246, xijie_jia_sri-surf_2016, wang_embedded_2014, kalms_accelerated_2017, lam_data-path_2019}) and GPUs (\cite{FuentesAlventosa2022GUDCannyAR, Ramkumar2019GPUAO, 9047171, Zhi2016RealizationOC, 6641456, 7490389, LI201978, Aniruddha2018ARI}) has become a popular research direction in both computer system and robotics. In recent years, computer system and robotics researchers have begun considering the hardware acceleration of feature detection within a visual SLAM pipeline (\cite{lentaris_hwsw_2016, chen_bucket-stream_2020, 10.1145/3558481.3591310, 8967814, 10546618, XUEHE2016104, liu_eslam_2019, xu_cnn-based_2020, liu_mobilesp_2022, piat_hwsw_2020, mandal_visual_2019, nikolic_synchronized_2014, tang_-soc_2018, wang_ac2slam_2021, gu_real-time_2022, 10.1109/IROS45743.2020.9340851}). However, none of these publications have considered a comparison of hardware-accelerated feature detectors integrated with a visual SLAM pipeline.

This paper presents the first comparative study on FPGA- and GPU-accelerated feature detectors considering a V-SLAM pipeline. ICE-BA is the selected V-SLAM pipeline due to its state-of-the-art accuracy, efficiency, consistency, and software modularity. This paper compares the GPU and FPGA implementations of FAST, Harris, and SuperPoint \cite{detone_superpoint_2018}. FAST and SuperPoint are selected because they represent state-of-the-art algorithmic and neural network-based feature detectors, respectively. Furthermore, FAST and SuperPoint provide repeatability \cite{4674368, detone_superpoint_2018} not offered by other non-learning-based detectors such as Harris, SIFT \cite{Lowe2004DistinctiveIF}, SUSAN \cite{Smith1997SUSANANA}, and Shi-Tomasi \cite{323794}. SuperPoint is selected over SiLK \cite{10378537}, as SuperPoint is a light CNN and therefore, more suitable for edge deployment \cite{xu_cnn-based_2020}. Harris is chosen for completeness.

The evaluation uses the Machine Hall (MH) sequences of the EuRoC data set and shows that when using FAST and Harris, the implementation reliant on GPU can achieve better run-time performance and energy efficiency than the FPGA implementation. However, when considering SuperPoint, its FPGA implementation can achieve better run-time performance and energy efficiency (up to 3.1$\times$ and 1.4$\times$ improvements, respectively) than the GPU implementation. The FPGA-accelerated ICE-BA can also achieve comparable run-time performance when compared to the GPU-accelerated ICE-BA, with better FPS in 2 out of 5 dataset sequences. However, when considering the accuracy, the results show that the GPU-accelerated ICE-BA is more accurate than the FPGA-accelerated ICE-BA in general. Furthermore, the use of hardware accelerators for feature detection could further improve the performance of the V-SLAM pipeline by having the global bundle adjustment module invoked less frequently, without sacrificing the accuracy.

\section{Background}

This section briefly introduces the background of FPGA, visual SLAM and feature detection.

\subsection{Field Programmable Gate Array}

An FPGA is an integrated circuit with a flexible hardware architecture that can be reconfigured after manufacture. An AMD FPGA SoC has a Processing System (PS) and Programmable Logic (PL). The PS typically consists of Arm cores, GPU, and DRAM. The PL consists of a matrix of programmable logic blocks connected by a programmable interconnect. Each programmable logic block has several Flip-Flops (FF) and Look-Up Tables (LUT). In addition to logic blocks, an FPGA also has several Digital Signal Processors (DSPs) and fast on-chip memory, commonly known as Block Memory (BRAM) and UltraRAM in AMD FPGAs, which are also connected via programmable interconnect. In addition to PS and PL, the latest AMD Versal FPGAs \cite{noauthor_versal_nodate} also feature AI Engines (AIE), which are dedicated to the acceleration of machine learning inference and signal processing.

%\subsubsection{AMD Vitis AI and Vitis Vision Library}

%The AMD Vitis Vision Library consists of a number of FPGA-accelerated computer vision kernels, such as FAST feature detector, Harris feature detector, image pyramid computation and Gaussian filter, developed using HLS C/C++. Vitis Vision Library allows software developers to utilize hardware-accelerated computer vision kernels in their design in a plug-and-play way. The AMD Vitis AI is a framework that can be leveraged to accelerate AI inference on AMD FPGAs. Vitis AI also provides optimized FPGA IPs for a wide range of models, such as MobileNet, ResNet, VGG, YOLO, BERT and SuperPoint.

\subsection{Visual SLAM}

Visual SLAM is a ubiquitous problem in areas such as the navigation of unmanned vehicles, VR, and AR. In general, visual SLAM can be thought of as the problem of building a map of the unknown environment using only visual sensors (e.g.\ stereo and monocular cameras), while determining the pose (position and orientation) of the agent within this environment at the same time.

The visual SLAM pipeline usually includes processing input from visual sensors to obtain observations of the environment and conduct probabilistic state estimation using the observations as constraints. Typically, a visual SLAM pipeline includes two main modules (i.e., localization and mapping modules), which are executed in parallel with synchronisation. The input from the visual sensors needs to be preprocessed first. After that, feature detection is conducted, followed by pose estimation within the localization module. Subsequently, the mapping module uses the keyframes to update the map and conducts local bundle adjustments, which jointly optimizes the recent poses and map points. In this case, the keyframes are data structures that contain the estimated pose and the features observed from that pose. The updated map can be used later by the localization module to conduct pose estimation. Loop closure detection is conducted when the incoming frame to the mapping module is a keyframe. When a loop is detected, global bundle adjustment will be invoked to close the loop in the trajectory. Global bundle adjustment only executes occasionally since it is a time-consuming process.

\subsubsection{ICE-BA}

ICE-BA is an efficient, sliding-window-based bundle adjustment solver for V-SLAMs. It can achieve better accuracy and robustness by leveraging a larger number of measurements, while at the same time being at least 10$\times$ more computationally efficient than other state-of-the-art implementations, such as OKVIS \cite{10.1177/0278364914554813}, iSAM2 \cite{5979641} and ORB-SLAM \cite{7219438}, by exploiting the sparseness of the matrix structure during optimization. Using a novel relative marginalization method, ICE-BA also improves global consistency. 

\subsection{Feature Detection}

\begin{algorithm}[h]
\small 
\SetAlgoLined
 \caption{A Generic Corner Detection Algorithm}
 \label{alg:generic_feature_detect}
 %\begin{algorithmic}[1]
  Pre-processing.\\
  \For{each scale}{
    \For{each pixel within the region of interest}{
         Search for pixel intensity change in the X and Y directions.\\
         Compute the Sum of Squared Differences (SSD) of the neighbour pixels around the potential corner.\\
         Shift the SSD of pixel intensity using Taylor expansion.\\
    }
     Post-processing with Non-Maxima Suppression (NMS).
 }
 %\end{algorithmic}
\end{algorithm}

A feature is a locally distinct pixel on an image, invariant to translation, rotation, and illumination. A generic corner detection algorithm is presented in Algorithm \ref{alg:generic_feature_detect}. The pre-processing step of corner detection includes colour conversion, blurring, and image pyramid building. Colour conversion converts an RGB image to grey-scale. The image is blurred by applying filters, such as the Gaussian filter, over the image. Real-world image is typically affected by noise, the Gaussian filter can smooth the image by filtering out the high-frequency noises. Feature detection is conducted with an image pyramid, to find features that are irrespective of scale changes. An image pyramid is a multi-scale representation of an image, which is typically generated by blurring and sub-sampling the original image. The higher the pyramid level, the fewer pixels are in the image. 

NMS is a common post-processing step in corner detection. NMS seeks to find the corner with the maximum score within a local region (region of interest) and remove other corners within the same region.

\subsubsection{FAST}

\begin{table}[]
\centering
\caption{FAST notations}
\begin{tabular}{|c|c|}
\hline
Symbols & Definitions \\ \hline
%\( Img \)      &    Image      \\ \hline
%\( W \)      &    Width of image \( Img \)      \\ \hline
%\( P \)        &    Pixel      \\ \hline
\( I_P \)        &  Intensity of pixel \( P \)       \\ \hline
%\( F \)        &   Feature       \\ \hline
\( t \)        &   FAST threshold       \\ \hline
\( S_F \)        &   FAST score of corner \( F \)       \\ \hline
$\varepsilon$        &   A small constant value       \\ \hline
\( N \)       &   Number of pixels processed in parallel      \\ \hline
%\( buf \)       &   A 2D array with a size of 7x(\( W \)/\( N \)) that buffers \( P \)      \\ \hline
\( buf \)       &   A 2D array that buffers pixel \( P \)      \\ \hline
%\( buf_s \)       &   A 2D array with a size of 7x\( N \)       \\ \hline
%\( buf_d \)       &   A 2D array with a size of 7x(\( N \)+7-1)      \\ \hline
%\( buf_{nms} \)       &   A 2D array with a size of 3x(\( W \)/\( N \)) that buffers \( P \)      \\ \hline
\( buf_{nms} \)       &   A 2D array that buffers pixel intensity \( I_P \)      \\ \hline
%\( buf_s_nms \)       &   A 2D array with a size of 3x\( N \)       \\ \hline
%\( buf_d_nms \)       &   A 2D array with a size of 3x(\( N \)+3-1)      \\ \hline
%\( diff \)       &   A 2D array with a size of Nx25 that stores the intensity difference between the centre \( P \) and every other \( P_i \) on the Bresenham circle    \\ \hline
%\( flags \)       &   A 2D array with a size of Nx25 that stores the intensity difference flag     \\ \hline
\( diff \)       &   \begin{tabular}[c]{@{}c@{}}A 2D array that stores the intensity \\differences between the centre pixel \\ \( P \) and pixels \( P_i \) on the Bresenham circle\end{tabular}    \\ \hline
\( flags \)       &   \begin{tabular}[c]{@{}c@{}}A 2D array that stores the intensity difference flag.\end{tabular}     \\ \hline
%RowInd       &   Row index of \( buf \)      \\ \hline
%RowIndNMS       &   Row index of \( buf_nms \)      \\ \hline
\( tag_P \)       &   A tag that determines whether pixel \( P \) is a feature or not     \\ \hline
\end{tabular}
\label{table:fast_notations}
\end{table}

\begin{algorithm}[h]
\small 
\SetAlgoLined
 \caption{FAST Feature Detector}
 \label{alg:fast}
    \textbf{Input: Image, \( t \)} \\
  \textbf{Output: A vector that contains information of detected features \( F \)} \\
  %\begin{algorithmic}[1]
  \For{each pixel \( P \) in the input image}{
    %\State Define the pixel intensity as \( I_P \)
    %\State Define the threshold as  \( T \)
    %\State Define the 16-pixel Bresenham circle whose radius is 3 with \( P \) as the centre, the intensity of each pixel on the Bresenham circle is represented as \( I_1 \), \( I_2 \),...,\( I_{16} \) 
  \For{each pixel \( P_i \) on the Bresenham circle}{
    \If{(\( I_{P_1} \) $>$ \( I_P \) + \( t \))$\cap$(\( I_{P_2} \) $>$ \( I_P \) + \( t \))$\cap$...$\cap$(\( I_{P_i} \) $>$ \( I_P \) + \( t \)) or (\( I_{P_1} \) $<$ \( I_P \) - \( t \))$\cap$(\( I_{P_2} \) $<$ \( I_P \) - \( t \))$\cap$...$\cap$(\( I_{P_i} \) $<$ \( I_P \) - \( t \))}{
         Pixel \( P \) is a corner
    }
  }
  %\State Define \(\varepsilon\) to be a small constant number
  \For{each detected corner \( F \)}{
     \While{\( F \) is still a corner}{
        %\State \( I_F \) = \( I_F \) + $\varepsilon$
         \( t \) = \( t \) + $\varepsilon$
     }
     %\State \( S_F \) = \( I_F \)
      \( S_F \) = \( t \)
  }
  %\State Define a 3*3 window centred around \( F \), each pixel surround \( F \) is represented as \( P_1 \), \( P_2 \),...,\( P_8 \)
  \For{each detected corner \( F \)}{
    %Non maxima suppression with a n*n window centred around \( F \)\\
    \For{each pixel \( P_n \) within the NMS window}{
        \If{(\( P_n \) is a corner) \(\cap\) (\( S_{P_n} \) $>$ \( S_F \))}{
             \( F \) is no longer a corner
             Break NMS loop
        }
    }
  }
 }
 %\end{algorithmic}
\end{algorithm}

FAST is a heuristic feature detector derived using an online learning method. Compared with other non-learning-based feature detectors such as SIFT, Harris, Shi-Tomasi, and SUSAN, FAST can achieve better repeatability and run-time performance.  Algorithm \ref{alg:fast} and Table \ref{table:fast_notations} summarize the FAST algorithm and the symbols it uses.

\subsubsection{Harris}

Harris is a local Sum of Squared Differences (SSD) based feature detector that is built on the Moravec detector \cite{10.5555/909315}. The Moravec detector defines features as pixels with low self-similarity in all directions. The self-similarity of an image patch can be measured by taking the SSD between an image patch and a shifted version of itself. The Harris detector is improved upon this by computing an approximation to the second derivative of the SSD with respect to the shift, which is more computationally efficient. This approximation could be calculated as,

\[M(x, y)=\sum_{u,v}\omega(u,v)\begin{bmatrix}
I_x^2 & I_x I_y \\
I_x I_y & I_y^2 
\end{bmatrix}\]

where \( I_x \) and \( I_y \) represent the derivative in x and y direction of the pixel intensity at pixel\((x+u, y+v)\) respectively. \(\omega(u,v)\) is the weighted averaging function.

Using \( M \), the Harris response \( R \) can be calculated as,

\[R=det(M)-k(trace(M))^2\]

where k is the weighting factor. A small Harris response represents a flat region, while a negative one represents an edge feature; a large positive Harris response represents a corner feature.

\subsubsection{SuperPoint}

SuperPoint is a self-supervised framework for training feature detectors and descriptors for multi-view geometry problems. It is an efficient, fully convolutional neural network that operates on full-sized images and can jointly compute pixel-level feature points and their descriptors in one forward pass. SuperPoint can achieve better repeatability than FAST, Harris, and Shi-Tomasi due to the proposed homographic adaptation, which is a multi-scale, multi-homography approach for improving feature detection repeatability. 

%could add a diagram for SP here...

\section{Related Works}

Ulusel \textit{et al}.\ \cite{ulusel_hardware_2016} have designed and compared hardware accelerators for FAST, BRIEF \cite{6081878} and BRISK \cite{6126542} using FPGA and GPU, respectively, whereas Kalms and Göhringer \cite{8056847} have designed and compared FPGA and GPU accelerators for AKAZE \cite{Alcantarilla2013FastED}. Possa \textit{et al}.\ \cite{possa_multi-resolution_2014} accelerated the Canny \cite{4767851} and Harris \cite{Harris1988ACC} feature detectors using FPGA and GPU and compared their run-time performance, power, and energy consumption. On the other hand, Qasaimeh \textit{et al}.\ \cite{8782524} have compared several computer vision kernels, including the Canny, FAST, and Harris feature detectors, from Nvidia’s VisionWorks library and Xilinx’s xfOpenCV library, using FPGA and GPU. Chouchene \textit{et al}.\ \cite{10.1504/IJAMC.2014.060506} implemented and compared the Sobel edge detector on CPU, GPU, and FPGA. Pauwels \textit{et al}.\ \cite{5936059} and Struyf \textit{et al}.\ \cite{struyf2014battle} both designed and compared hardware accelerators for Gabor using GPU and FPGA. Table \ref{table:contribution} summarizes the related work.

However, these works only compared the FPGA- and GPU-accelerated feature detectors as standalone components, without considering a full visual SLAM pipeline. Such a comparison can be considered incomplete since the feature detector is usually part of other applications, such as image classification, object detection, and the front-end of visual SLAM. Furthermore, none of the previous work compares non-learning-based against learning-based feature detectors on GPU and FPGA architectures.

\begin{table}[t]
\centering
\caption{Summary of comparative studies on GPU- and FPGA-accelerated feature detectors and the contribution of this paper}
\begin{tabular}{ |c|c|c|c| } 
\hline
& \begin{tabular}[c]{@{}c@{}}Feature Detection \\Algorithms\end{tabular} & \begin{tabular}[c]{@{}c@{}}SLAM\\ Integration\end{tabular}  \\ \hline
%\cite{lam_data-path_2019} & X & X & X \\ 
%\hline
%\cite{brenot_fpga_2015} & X & X & X \\ 
%\hline
%\cite{ding_multi-scale_2018} & X & X & X \\ 
%\hline
\cite{ulusel_hardware_2016} & FAST, BRIEF, BRISK  & \xmark   \\ 
\hline
%\cite{nikolic_synchronized_2014} & X & \checkmark & X \\ 
%\hline
%\cite{kalms_full-hd_2018} & \checkmark & X & X \\ 
%\hline
%\cite{mandal_visual_2019} & X & \checkmark & X \\ 
%\hline
%\cite{lentaris_hwsw_2016} & X & \checkmark & X \\ 
%\hline
%\cite{piat_hwsw_2020} & X & \checkmark & X \\ 
%\hline
%\cite{piat_hwsw_2020} & X & \checkmark & X \\ 
%\hline
\cite{8056847} & AKAZE & \xmark   \\ 
\hline
\cite{possa_multi-resolution_2014} & Canny, Harris & \xmark   \\ 
\hline
\cite{8782524} & Canny, FAST, Harris & \xmark   \\ 
\hline
\cite{5936059} & Gabor & \xmark   \\ 
\hline
\cite{struyf2014battle} & Gabor & \xmark   \\ 
\hline
\cite{10.1504/IJAMC.2014.060506} & Sobel & \xmark   \\ 
\hline
%\cite{ABOUZAHIR201814} & X & \checkmark & \checkmark \\ 
%\hline
%\cite{zhou_guidance_2015} & \checkmark & \checkmark & X \\ 
%\hline
%\cite{nikolic_synchronized_2014} & \checkmark & \checkmark & X \\ 
%\hline
\textbf{Ours} & FAST, Harris, SuperPoint & \textbf{\checkmark}  \\ 
\hline
\end{tabular}
\label{table:contribution}
\end{table} 

\section{Experiments}
\label{sec:Methodology}
This section introduces the experimental methodology and setup.
\subsection{Hardware and Software Setup}

\begin{table*}[t]
\centering
\caption{Summary of hardware and software specifications and configurations.}
\begin{threeparttable}
\begin{tabular}{ |c|c|c|c| } 
\hline
& Intel Xeon &\begin{tabular}[c]{@{}c@{}}Nvidia Jetson AGX Orin\end{tabular} & \begin{tabular}[c]{@{}c@{}}AMD Versal VCK190 \end{tabular}\\ \hline
Processor & \begin{tabular}[c]{@{}c@{}}Intel Xeon W-2123 (8 cores, 3.6 GHz)\end{tabular}  &\begin{tabular}[c]{@{}c@{}}Arm Cortex-A78AE (12 cores, 2.2 GHz, \\64 KB L1i\footnotemark and L1d\footnotemark, 3 MB L2, 6 MB L3)\end{tabular} & \begin{tabular}[c]{@{}c@{}}Arm Cortex-A72 (2 cores, 1.2 GHz, \\48 KB L1i, 32 KB L1d, 1 MB L2)\end{tabular} \\ 
\hline
Accelerator & N/A  &\begin{tabular}[c]{@{}c@{}}Nvidia Ampere GPU \\(2048 CUDA cores, 64 Tensor cores, 1.3 GHz)\end{tabular} & \begin{tabular}[c]{@{}c@{}}Vitis Vision FAST, 150 MHz, \\Vitis AI SuperPoint, 1.3 GHz\end{tabular} \\ 
\hline
DRAM & 64 GB DDR4 &\begin{tabular}[c]{@{}c@{}}32 GB LPDDR5 \end{tabular} & \begin{tabular}[c]{@{}c@{}}8 GB DDR4\end{tabular}\\ 
\hline
OS & Ubuntu 20.04 &\begin{tabular}[c]{@{}c@{}}Ubuntu 20.04\end{tabular} & \begin{tabular}[c]{@{}c@{}}Petalinux 2023.2\end{tabular} \\ 
\hline
Compiler & GCC-9.4 &GCC-9.4, NVCC & \begin{tabular}[c]{@{}c@{}}GCC-9.3, Vitis HLS 2023.2\end{tabular} \\ 
\hline
\begin{tabular}[c]{@{}c@{}}Compiler\\ Flags\end{tabular} & \multicolumn{3}{c|}{\begin{tabular}[c]{@{}c@{}}\texttt{-O3}, \texttt{-fno-math-errno}, \texttt{-funroll-loops}, \texttt{-fno-finite-math-only}\end{tabular}} \\ 
\hline
\begin{tabular}[c]{@{}c@{}}ISA\end{tabular} &\begin{tabular}[c]{@{}c@{}}Intel AVX \& SSE\end{tabular} &\multicolumn{2}{c|}{\begin{tabular}[c]{@{}c@{}}Arm NEON\end{tabular}} \\ 
\hline
Library & OpenCV 4.5.5 &\begin{tabular}[c]{@{}c@{}}OpenCV 4.5.4, CUDA 11.4, TensorRT 8.4\end{tabular} & \begin{tabular}[c]{@{}c@{}}OpenCV 4.5.5, Vitis Vision 2022.1, \\Vitis AI 3.0\end{tabular} \\ 
\hline
\end{tabular}
\begin{tablenotes}
\item[1] \footnotesize{L1i stands for L1 instruction cache.}
\item[2] \footnotesize{L1d stands for L1 data cache.}
\end{tablenotes}
\end{threeparttable}
\label{table:setup}
\end{table*}

Table \ref{table:setup} summarizes the hardware and software setup of the experiments. The evaluation uses two state-of-the-art energy-efficient SoCs, the Nvidia Jetson AGX Orin, and the AMD Versal VCK190. For completeness, an Intel-based workstation is also included. 

\subsection{Datasets}

The evaluation uses the Machine Hall (MH) sequences from the EuRoC data set \cite{doi:10.1177/0278364915620033}, which has a resolution of 752$\times$480. Each sequence is categorized into either "easy", "medium", or "difficult". The environment in "easy" sequences has good texture and illumination, whereas "medium" sequences contain fast motion and bright scenes. "Difficult" sequences have scenes with fast motion and poor illumination. The images of the MH sequences are captured by a stereo camera at 20 Hz, while the IMU data is synchronized at 200 Hz. Only the images from the left camera and IMU data are used in the experiments. 

\subsection{Evaluation}

\begin{figure}[tb]
    \centering
    \includegraphics[width=1.0\linewidth]{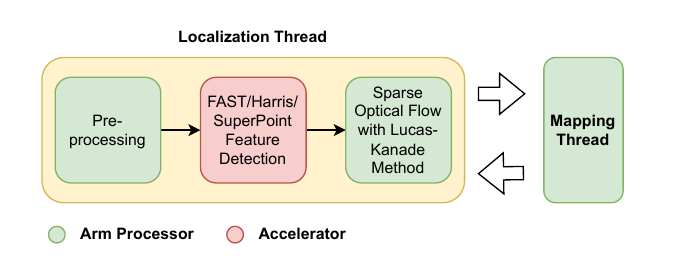}
    \caption{High-level overview of the hardware-accelerated ICE-BA pipeline. Best viewed in color.}
    \label{fig:high_level_sys}
\end{figure}

Figure \ref{fig:high_level_sys} presents a high-level overview of the accelerated ICE-BA pipeline. Modules colored in green are executed on the Arm processors of the considered SoCs. The module colored in red (feature detector) is accelerated on the embedded GPU or FPGA. On the Versal VCK190, the FAST and Harris FPGA accelerators, from the Vitis Vision Library \cite{noauthor_vitis_nodate-1}, are implemented on programmable logic, whereas the SuperPoint FPGA accelerator, from the Vitis AI Library \cite{vitis_ai}, is implemented using both programmable logic and the AMD AI Engine. On the Nvidia Jetson Orin, the SuperPoint implementation from \cite{xu2023airvo} is accelerated on the embedded GPU with TensorRT, whereas the recent Faster than FAST (FTFast) \cite{10.1109/IROS45743.2020.9340851} is selected as the most optimized GPU implementation for FAST. For the Harris GPU accelerator, we selected the implementation from the Nvidia VPI library with the CUDA backend. An Intel system is also utilized to provide software baselines using FAST and Harris implementations from OpenCV.

\begin{table}[tb]
\centering
\caption{The algorithmic parameters of FAST and Harris}
\begin{tabular}{|c|cc|}
\hline
Parameters                                                                  & \multicolumn{1}{c|}{FAST}                                                         & Harris \\ \hline
FAST Type                                                                   & \multicolumn{1}{c|}{FAST-9}                                                       & N/A    \\ \hline
Threshold                                                                   & \multicolumn{2}{c|}{10}                                                                    \\ \hline
Sensitivity \( K \)                                                               & \multicolumn{1}{c|}{N/A}                                                          & 0.04   \\ \hline
NMS?                                                                        & \multicolumn{2}{c|}{Yes}                                                                   \\ \hline
NMS Window Size                                                             & \multicolumn{1}{c|}{3$\times$3}                                                          & 2$\times$2    \\ \hline
\begin{tabular}[c]{@{}c@{}}Feature Score \\ Computation Method\end{tabular} & \multicolumn{1}{c|}{\begin{tabular}[c]{@{}c@{}}Maximum \\ Threshold\end{tabular}} & N/A    \\ \hline
Sobel Filter Size                                                           & \multicolumn{1}{c|}{N/A}                                                          & 7$\times$7    \\ \hline
Neighbor Block Size                                                         & \multicolumn{1}{c|}{N/A}                                                          & 7$\times$7   \\ \hline
\end{tabular}
\label{table:fast_harris_config}
\end{table}

Both SuperPoint models from \cite{vitis_ai} and \cite{xu2023airvo} are pre-trained using the MS COCO 2014 dataset \cite{Lin2014MicrosoftCC}. We did not limit the number of feature points that can be detected by the FAST, Harris, and SuperPoint accelerators. Each MH sequence is pre-loaded completely into DRAM before the processing starts. We execute ICE-BA in its monocular mode. In the ICE-BA mapping thread, the local and global bundle adjustments are executed in parallel on two different processor threads. Table \ref{table:fast_harris_config} summarizes the algorithmic parameters of the FAST and Harris feature detectors. The hardware accelerators share the same algorithmic parameters as the software baseline. 

\begin{table}[tb]
\centering
\caption{Configurations of Nvidia AGX Orin}
\begin{tabular}{ |c|c|c| } 
\hline
Configuration &\begin{tabular}[c]{@{}c@{}}Orin Versal \\Similar (VS)\end{tabular} & Orin max\\ \hline
\# of Processor Cores &2 & 12 \\ 
\hline
Processor Clock Frequency &600 MHz & 2.2 GHZ \\ 
\hline
GPU Clock Frequency & \begin{tabular}[c]{@{}c@{}}150 MHz (FTFast \\and VPI Harris) /\\1.3 GHz (SuperPoint)\end{tabular}  & 1.3 GHz\\ 
\hline
\end{tabular}
\label{table:orin_config}
\end{table}

 Table \ref{table:orin_config} summarizes the configurations of the Nvidia Jetson Orin. We used two different configurations of the Orin during the evaluation. The Orin max configuration enables maximum performance by enabling all twelve Arm cores, and fixing the clock frequency of the Arm cores and the GPU to 2.2 GHz and 1.3 GHz, respectively. The number of online Arm processor cores on Orin is configured by modifying the Linux kernel files, to bring several selected Arm cores online/offline. The clock frequency of the Arm processors and the GPU is fixed by disabling Dynamic Voltage and Frequency Scaling (DVFS) and modifying the Linux kernel files to apply the frequency we need. 
 
 The Orin Versal Similar (VS) configuration aims to approximate the processing cores and accelerator configuration of the Versal VCK190 on the Orin platform, for a fair comparison. This is because the Orin has 12 Arm A78 cores at 2.2 GHz, whereas the VCK190 only has 2 Arm A72 cores at 1.2 GHz. Note that the Arm A78 has a more advanced microarchitecture than the Arm A72 (see Table \ref{table:setup}). The following experiments are conducted to determine the Arm processor configuration. ICE-BA is executed on VCK190 and Orin using only two Arm cores and FAST implementation from OpenCV, the run-time of its localization thread is measured. ICE-BA is executed using different clock frequencies of the Arm processors. We use frequencies available with both the AMD Petalinux 2023.2 OS and the Ubuntu 20.04 OS on Orin, which are 300 MHz, 400 MHz, 600 MHz, and 1.2 GHz. The dataset we used in this experiment is the MH01 sequence. 

\begin{table}[tbh!]
\centering
\caption{Run-time (ms) of the ICE-BA localization thread.}
\begin{tabular}{|c|c|c|}
\hline
\begin{tabular}[c]{@{}c@{}}Clock \\Frequency (MHz)\end{tabular} & \begin{tabular}[c]{@{}c@{}}Arm A72\\ (max 1.2 GHz)\end{tabular} &  \begin{tabular}[c]{@{}c@{}}Arm A78\\ (max 2.2 GHz)\end{tabular}\\ \hline
300                                                & 46.3 & 34                                                                \\ \hline
400                                                & 46.3 & 20.6                                                                \\ \hline
600  & 31  & 14.9                                                                \\ \hline
1200 & 15.8 & 7.4                                                               \\ \hline
2200 & N/A & 4.4 \\ \hline
\end{tabular}
\label{table:vs_compare}
\end{table}
 
 Table \ref{table:vs_compare} illustrates the different run-time achieved by the ICE-BA localization thread under different clock frequencies with 2 Arm A72/A78 cores. Note for the 600 MHz frequency, the Arm Cortex A78 cores deliver 14.9 ms, while the Arm Cortex A72 cores deliver 15.8 ms. Thus, for the Orin VS configuration, the number of power-on Arm cores is 2. The clock frequency of the two Arm cores is 600 MHz. The GPU clock frequency is 150 MHz under this configuration when executing FTFast, to be in line with the frequency of the FAST FPGA accelerator.

\section{Experimental Results and Analysis}
\label{sec:evaluation}
This section presents the run-time performance, power, and energy efficiency results of the FPGA- and GPU-accelerated FAST, Harris, and SuperPoint. The run-time performance, accuracy, power, and energy efficiency results of ICE-BA integrated with different hardware-accelerated feature detectors are also presented.

\subsection{Results of the Feature Detector Hardware Accelerators} 

\begin{table}[]
\centering
\caption{Power comparison among GPU- and FPGA-accelerated feature detectors and the software baseline}
\begin{threeparttable}
\begin{tabular}{|c|c|ccc|}
\hline
\multirow{2}{*}{\begin{tabular}[c]{@{}c@{}}Feature \\ Detectors\end{tabular}} & \multirow{2}{*}{\begin{tabular}[c]{@{}c@{}}Implementations \\ and System Configs\end{tabular}} & \multicolumn{3}{c|}{Power (W)}                                            \\ \cline{3-5} 
                                                                             &                                                                                                & \multicolumn{1}{c|}{Processor} & \multicolumn{1}{c|}{Accelerator} & Total \\ \hline
\multirow{4}{*}{FAST}                                                        & OpenCV + Xeon                                                                                  & \multicolumn{1}{c|}{120\footnotemark}       & \multicolumn{1}{c|}{N/A}         & 120   \\ \cline{2-5} 
                                                                             & FTFast + Orin VS                                                                               & \multicolumn{1}{c|}{0.8}       & \multicolumn{1}{c|}{5.6}         & 12.6  \\ \cline{2-5} 
                                                                             & FTFast + Orin max                                                                              & \multicolumn{1}{c|}{4.4}       & \multicolumn{1}{c|}{8.8}         & 20.4  \\ \cline{2-5} 
                                                                             %& \begin{tabular}[c]{@{}c@{}}Vitis FAST AO \\ + VCK190\end{tabular}                              & \multicolumn{1}{c|}{4.9}       & \multicolumn{1}{c|}{9.3}         & 15.2  \\ \cline{2-5} 
                                                                             & \begin{tabular}[c]{@{}c@{}}Vitis FAST \end{tabular}                              & \multicolumn{1}{c|}{4.9}       & \multicolumn{1}{c|}{10.8}        & 16.7  \\ \hline
\multirow{4}{*}{Harris}                                                        & OpenCV + Xeon                                                                                  & \multicolumn{1}{c|}{120}       & \multicolumn{1}{c|}{N/A}         & 120   \\ \cline{2-5} 
                                                                             & VPI + Orin VS                                                                               & \multicolumn{1}{c|}{3.8}       & \multicolumn{1}{c|}{6}         & 17.5  \\ \cline{2-5} 
                                                                             & VPI + Orin max                                                                              & \multicolumn{1}{c|}{4.4}       & \multicolumn{1}{c|}{8.8}         & 21.5  \\ \cline{2-5} 
                                                                             %& \begin{tabular}[c]{@{}c@{}}Vitis FAST AO \\ + VCK190\end{tabular}                              & \multicolumn{1}{c|}{4.9}       & \multicolumn{1}{c|}{9.3}         & 15.2  \\ \cline{2-5} 
                                                                             & \begin{tabular}[c]{@{}c@{}}Vitis Harris \end{tabular}                              & \multicolumn{1}{c|}{4.6}       & \multicolumn{1}{c|}{10.3}        & 16.5  \\ \hline
\multirow{3}{*}{SuperPoint}                                                  & Orin VS                                                                                        & \multicolumn{1}{c|}{2.4}       & \multicolumn{1}{c|}{8.8}         & 18.5  \\ \cline{2-5} 
                                                                             & Orin max                                                                                       & \multicolumn{1}{c|}{4.4}       & \multicolumn{1}{c|}{9.6}         & 22    \\ \cline{2-5} 
                                                                             & \begin{tabular}[c]{@{}c@{}}Vitis SuperPoint\end{tabular}                           & \multicolumn{1}{c|}{5.2}       & \multicolumn{1}{c|}{21.1}        & 30    \\ \hline
\end{tabular}
\begin{tablenotes}
\item[3] \footnotesize{The TDP (Thermal Design Power) of the Intel Xeon W-2123 system.}
\end{tablenotes}
\end{threeparttable}
\label{table:power}
\end{table}

\begin{figure*}[tb]
    \centering
    \includegraphics[width=1.0\linewidth]{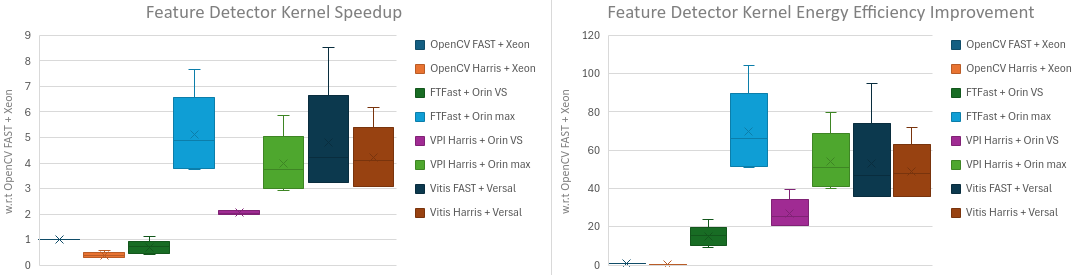}
    \caption{Speedups and energy efficiency improvements of different FAST and Harris accelerators w.r.t OpenCV FAST + Xeon. Best viewed in color.}
    \label{fig:accel_speedup_energy}
\end{figure*}

\begin{figure*}[tb]
    \centering
    \includegraphics[width=1.0\linewidth]{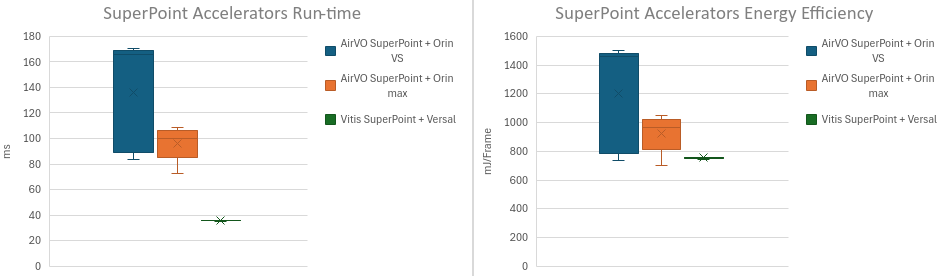}
    \caption{Run-time and energy efficiency of different SuperPoint accelerators. Best viewed in color.}
    \label{fig:sp_accel_perf_energy}
\end{figure*}

Figure \ref{fig:accel_speedup_energy} summarizes the results of speedup and energy efficiency improvement of the GPU- and FPGA-accelerated FAST and Harris, with respect to the FAST software baseline on Xeon. 

Regarding the FAST accelerators, FTFast+Orin max achieves the best run-time performance in 4 out of 5 sequences (except for MH01), and the best energy efficiency (only 2.2 - 2.3 mJ/Frame) across all MH sequences, followed by the Vitis FAST accelerator on VCK190, which is 12.8\% - 15.2\% slower. However, note that the GPU clock frequency is 1.3 GHz under the Orin max configuration, whereas the clock frequency of the FAST FPGA accelerator is 150 MHz. %Furthermore, the Vitis FAST accelerator is processing 8 pixels in one clock cycle, whereas FTFast uses 128 CUDA threads per thread block. 
Compared with FTFast+Orin VS, where the GPU clock frequency is the same as the FPGA accelerators, the Vitis FAST accelerator can achieve 5.9$\times$ - 7.7$\times$ speedups and 3.3$\times$ - 4.3$\times$ improvements in energy efficiency. The FAST FPGA accelerator yields better run-time performance than FTFast+Orin VS because it prioritizes latency over scalability by processing 8 pixels per clock cycle. Furthermore, the Vitis FAST implementation utilizes approximation, reduced precision, and function overlapping. The Vitis FAST accelerator uses shift operations, which are less computationally expensive and require fewer hardware resources to implement, to approximate multiplication and division operations. Furthermore, the NMS function overlaps with the FAST score computation function, and fixed-point numbers are used instead of floating-point numbers. Compared with the OpenCV software baseline on an Intel Xeon processor, FTFast+Orin max achieves 3.7$\times$ - 7.6$\times$ speedups and 52$\times$ - 104$\times$ improvements in energy efficiency, while the Vitis FAST accelerator achieves 5.1$\times$ - 8.5$\times$ speedups and 38.3$\times$ - 95.6$\times$ improvements in energy efficiency.

In terms of the Harris accelerators, the Vitis Harris accelerator achieves the best run-time performance across all MH sequences, with 10.6$\times$ - 11$\times$ speedups against the OpenCV baseline on an Intel Xeon processor, and 1.01$\times$ - 1.1$\times$ speedups against the VPI Harris+Orin max. The Vitis Harris accelerator is slightly faster than the VPI Harris+Orin max due to the use of reduced precision numbers. On the other hand, VPI Harris+Orin max achieves the best energy efficiency across all MH sequences, with 136$\times$ - 146$\times$ improvements against the software baseline on the Intel Xeon, and 1.02$\times$ - 1.1$\times$ improvements against the Vitis Harris accelerator.

Figure \ref{fig:sp_accel_perf_energy} summarizes the run-time and energy efficiency of different SuperPoint GPU and FPGA accelerators. The Vitis SuperPoint accelerator with VCK190 achieves the best run-time performance (28 FPS) and energy efficiency (except for MH04, 745 - 758 mJ/Frame) across all MH sequences, with 2$\times$ - 3.1$\times$ speedups and 1.2$\times$ - 1.4$\times$ improvements in energy efficiency, compared with SuperPoint+Orin max. Note that the Vitis SuperPoint accelerator is the only accelerator that can achieve real-time performance, whereas SuperPoint+Orin max can only yield up to 14 FPS. The Vitis SuperPoint accelerator can achieve better run-time performance because it is quantized to use INT8 precision, whereas the SuperPoint accelerator from \cite{xu2023airvo} is quantized with FP16 precision. 

Compared with the FAST accelerators, Harris accelerators on both hardware platforms exhibit worse runtime and energy efficiency, especially for the GPU implementations. This is because FAST is a more efficient algorithm than Harris, as demonstrated in \cite{4674368}. Furthermore, Harris accelerators have similar power consumption to the FAST accelerators, as reported in Table \ref{table:power}.

Compared with the FAST and Harris accelerators, SuperPoint accelerators on both hardware platforms exhibit worse run-time and energy efficiency. This is expected since SuperPoint is more computationally expensive than FAST and Harris. Further, according to Table \ref{table:power}, the SuperPoint accelerators have a higher power consumption than both the FAST and Harris accelerators, especially for FPGA (21.1 W vs 10.8 W vs 10.3 W). This is because the power of an FPGA accelerator is proportional to its clock frequency and the area it occupies. The AMD Deep Learning Processor Unit (DPU) on which the Vitis SuperPoint executes, operates at a higher frequency (1.3 GHz vs 150 MHz) and occupies more area (FF: 28\% vs 0.82\% vs 0.97\%, LUT: 45\% vs 2.91\% vs 2.01\%, DSP: 42\% vs 0\% vs 0\%, BRAM: 73\% vs 1.24\% vs 3.62\%, AIE: 48\% vs 0\% vs 0\%) than the Vitis FAST and Harris accelerators.

\subsection{Results of the Hardware-accelerated ICE-BA} 

\begin{figure*}[tb]
    \centering
    \includegraphics[width=1.0\linewidth]{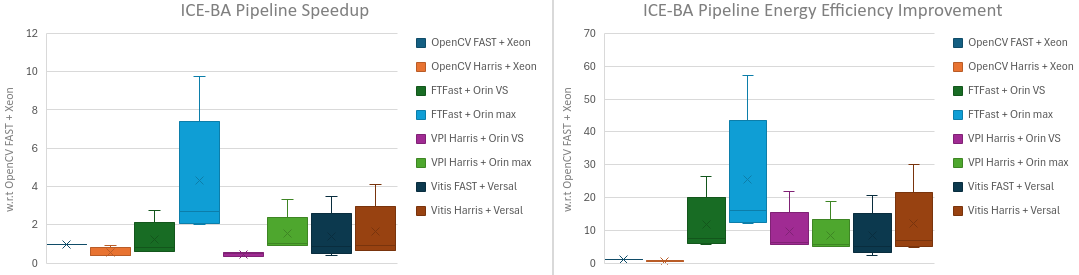}
    \caption{Speedups and energy efficiency improvements of ICE-BA integrated with different FAST and Harris accelerators w.r.t ICE-BA + OpenCV FAST + Xeon. Best viewed in color.}
    \label{fig:iceba_speedup_energy}
\end{figure*}

\begin{figure*}[tb]
    \centering
    \includegraphics[width=1.0\linewidth]{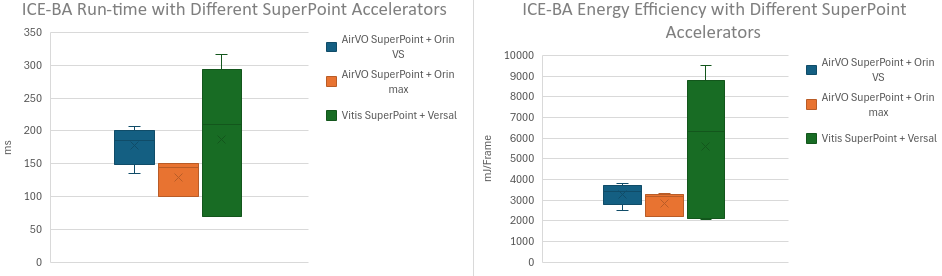}
    \caption{Run-time and energy efficiency of ICE-BA integrated with different SuperPoint accelerators. Best viewed in color.}
    \label{fig:iceba_sp_perf_energy}
\end{figure*}

\begin{figure}[tb]
    \centering
    \includegraphics[width=1.0\linewidth]{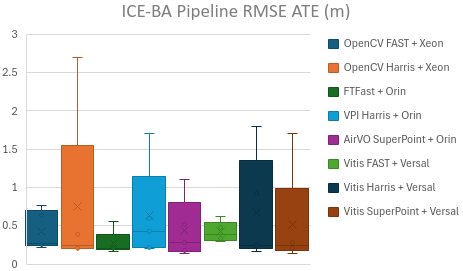}
    \caption{Accuracy of ICE-BA integrated with different FAST, Harris, and SuperPoint accelerators. Best viewed in color.}
    \label{fig:iceba_accuracy}
\end{figure}

% \usepackage{multirow}
\begin{table*}[]
\centering
\caption{Run-time performance of FAST GPU and FPGA accelerators, ICE-BA localization thread and pipeline, as well as ICE-BA accuracy across all MH sequences}
\begin{tabular}{|c|c|ccc|c|}
\hline
\multirow{2}{*}{Dataset Sequences} & \multirow{2}{*}{Implementations and System Configs} & \multicolumn{3}{c|}{Run-time (ms)}                                                          & \multirow{2}{*}{RMSE ATE (m)} \\ \cline{3-5}
                                   &                                                     & \multicolumn{1}{c|}{Feature Detector} & \multicolumn{1}{c|}{Localization Thread} & Pipeline &                               \\ \hline
\multirow{4}{*}{MH01}              & OpenCV FAST + Orin max                                 & \multicolumn{1}{c|}{2.41}             & \multicolumn{1}{c|}{4.03}                & 30.69    & 0.26                          \\ \cline{2-6} 
                                   & FTFast + Orin max                                      & \multicolumn{1}{c|}{0.26}             & \multicolumn{1}{c|}{2.84}                & 12.77    & 0.22                          \\ \cline{2-6} 
                                   & OpenCV FAST + Versal                                & \multicolumn{1}{c|}{7.35}             & \multicolumn{1}{c|}{15.82}               & 128.19   & 0.23                          \\ \cline{2-6} 
                                   & Vitis FAST                                          & \multicolumn{1}{c|}{0.23}             & \multicolumn{1}{c|}{10.96}               & 42.07    & 0.34                          \\ \hline
\multirow{4}{*}{MH02}              & OpenCV FAST + Orin max                                 & \multicolumn{1}{c|}{1.88}             & \multicolumn{1}{c|}{4.03}                & 31.19    & 0.27                          \\ \cline{2-6} 
                                   & FTFast + Orin max                                      & \multicolumn{1}{c|}{0.26}             & \multicolumn{1}{c|}{2.9}                 & 13.19    & 0.16                          \\ \cline{2-6} 
                                   & OpenCV FAST + Versal                                & \multicolumn{1}{c|}{7.15}             & \multicolumn{1}{c|}{15.61}               & 128.63   & 0.29                          \\ \cline{2-6} 
                                   & Vitis FAST                                          & \multicolumn{1}{c|}{0.3}              & \multicolumn{1}{c|}{11.19}               & 64.72    & 0.29                          \\ \hline
\multirow{4}{*}{MH03}              & OpenCV FAST + Orin max                                 & \multicolumn{1}{c|}{1.66}             & \multicolumn{1}{c|}{3.64}                & 24.26    & 0.22                          \\ \cline{2-6} 
                                   & FTFast + Orin max                                      & \multicolumn{1}{c|}{0.26}             & \multicolumn{1}{c|}{2.58}                & 8.97     & 0.21                          \\ \cline{2-6} 
                                   & OpenCV FAST + Versal                                & \multicolumn{1}{c|}{6.34}             & \multicolumn{1}{c|}{14.47}               & 96.12    & 0.15                          \\ \cline{2-6} 
                                   & Vitis FAST                                          & \multicolumn{1}{c|}{0.3}              & \multicolumn{1}{c|}{9.54}                & 28.16    & 0.62                          \\ \hline
\multirow{4}{*}{MH04}              & OpenCV FAST + Orin max                                 & \multicolumn{1}{c|}{1.27}             & \multicolumn{1}{c|}{3.39}                & 50.68    & 0.46                          \\ \cline{2-6} 
                                   & FTFast + Orin max                                      & \multicolumn{1}{c|}{0.26}             & \multicolumn{1}{c|}{2.51}                & 9.97     & 0.21                          \\ \cline{2-6} 
                                   & OpenCV FAST + Versal                                & \multicolumn{1}{c|}{5.07}             & \multicolumn{1}{c|}{13.44}               & 157.78   & 0.36                          \\ \cline{2-6} 
                                   & Vitis FAST                                          & \multicolumn{1}{c|}{0.29}             & \multicolumn{1}{c|}{9.09}                & 29.92    & 0.39                          \\ \hline
\multirow{4}{*}{MH05}              & OpenCV FAST + Orin max                                 & \multicolumn{1}{c|}{1.26}             & \multicolumn{1}{c|}{3.22}                & 79.28    & 0.82                          \\ \cline{2-6} 
                                   & FTFast + Orin max                                      & \multicolumn{1}{c|}{0.25}             & \multicolumn{1}{c|}{2.61}                & 15.94    & 0.56                          \\ \cline{2-6} 
                                   & OpenCV FAST + Versal                                & \multicolumn{1}{c|}{5.02}             & \multicolumn{1}{c|}{12.9}                & 318.93   & 1.1                           \\ \cline{2-6} 
                                   & Vitis FAST                                          & \multicolumn{1}{c|}{0.3}              & \multicolumn{1}{c|}{9.5}                 & 44.47    & 0.48                          \\ \hline
\end{tabular}
\label{table:iceba_frontend_backend_fast}
\end{table*}

Figure \ref{fig:iceba_speedup_energy} summarizes the speedup and energy efficiency improvements of the ICE-BA pipeline integrated with GPU- and FPGA-accelerated FAST and Harris, with respect to the software baseline (ICE-BA+OpenCV FAST+Xeon). Figure \ref{fig:iceba_accuracy} summarizes the accuracy (in RMSE ATE) of the ICE-BA pipeline integrated with different FAST, Harris, and SuperPoint accelerators.

Regarding ICE-BA integrated with the FAST accelerators, ICE-BA+FTFast+Orin max achieves the best run-time performance and energy efficiency across all MH sequences. The run-time performance and energy efficiency of the pipeline can be as low as 9 ms (111 FPS) and 183 mJ/Frame, respectively. ICE-BA integrated with the Vitis FAST accelerator yields worse performance and therefore worse energy efficiency, due to the disadvantages in processor microarchitecture (Arm A72 vs Arm A78), the number of processor cores (2 vs 12), and the clock frequency (1.2 GHz vs 2.2 GHz), between the Orin and VCK190. However, compared with ICE-BA+FTFast+Orin VS, ICE-BA with the Vitis FAST accelerator can achieve comparable run-time performance, with slightly better performance in the MH03, MH04, and MH05 sequences. However, ICE-BA with the FAST FPGA accelerator yields worse energy efficiency compared to ICE-BA+FTFast+Orin VS, due to higher power consumption (12.6 W vs 16.7 W, see Table \ref{table:power}). Compared with the software baseline on Xeon, ICE-BA+FTFast+Orin max pipeline achieves 2.1$\times$ - 10.5$\times$ speedups and 11.9$\times$ - 57.3$\times$ improvements in energy efficiency. ICE-BA pipeline integrated with the Vitis FAST accelerator achieves 3$\times$ - 25.1$\times$ improvements in energy efficiency when compared to the software baseline. Regarding accuracy, in general, ICE-BA+FTFast yields slightly better accuracy than the software baseline, whereas ICE-BA integrated with the Vitis FAST accelerator exhibits worse accuracy than ICE-BA+FTFast, except for MH05. This is mainly due to the use of approximation and reduced-precision numbers, where shift operations are used to approximate multiplication and division operations, and fixed-point numbers are used instead of floating-point numbers. 

In terms of ICE-BA integrated with the Harris accelerators, ICE-BA+VPI Harris+Orin max achieves the best run-time performance in the MH01 and MH02 sequences, whereas the ICE-BA pipeline integrated with the Vitis Harris accelerator achieves the best run-time performance in the MH03, MH04, and MH05 sequences. It is surprising to see ICE-BA+Vitis Harris demonstrates better run-time performance than ICE-BA+VPI Harris+Orin max in "medium" and "difficult" dataset sequences with fast motion and poor illumination, while having a disadvantage in the Arm processor microarchitecture. In terms of energy efficiency, ICE-BA+VPI Harris+Orin VS is the most energy efficient one in the MH01 and MH02 sequences, while ICE-BA+Vitis Harris is the most energy efficient implementation in the MH03, MH04, and MH05 sequences. With regards to accuracy, ICE-BA integrated with the Harris FPGA accelerator is more accurate than the GPU counterpart in "easy" sequences (MH01 and MH02), whereas ICE-BA integrated with the Harris GPU accelerator is more accurate in "medium" and "difficult" sequences (MH03, MH04, and MH05). Compared with the software baseline on Xeon, ICE-BA+VPI Harris+Orin max pipeline achieves 2.2$\times$ - 3.6$\times$ speedups and 12.2$\times$ - 20$\times$ improvements in energy efficiency. ICE-BA pipeline integrated with the Vitis Harris accelerator achieves 1.7$\times$ - 4.4$\times$ speedups and 12.6$\times$ - 33.4$\times$ improvements in energy efficiency when compared to the software baseline.

Figure \ref{fig:iceba_speedup_energy} summarizes the run-time performance and energy efficiency of the ICE-BA pipeline integrated with GPU- and FPGA-accelerated SuperPoint. It is interesting to see that ICE-BA+Vitis SuperPoint can achieve the best run-time performance and energy efficiency with sequences MH01 and MH02, despite the limited Arm cores and their frequency on the VCK190. We believe this is because MH01 and MH02 are "easy" sequences that represent scenes with good textures. Compared with ICE-BA+SuperPoint+Orin max, ICE-BA+Vitis SuperPoint achieves up to 1.5$\times$ speedups and 1.1$\times$ improvements in energy efficiency with MH01 and MH02 sequences. ICE-BA+Vitis SuperPoint can also achieve comparable run-time performance with ICE-BA+SuperPoint+Orin max in the rest of the sequences. ICE-BA+SuperPoint+Orin max yields the best run-time performance (up to 7 FPS) and energy efficiency with sequences MH03, MH04, and MH05. In terms of accuracy, ICE-BA integrated with the SuperPoint GPU accelerator is more accurate than ICE-BA+Vitis SuperPoint in general, except for MH04. ICE-BA+Vitis SuperPoint is less accurate since the Vitis SuperPoint is quantized using INT8 precision, while the SuperPoint GPU accelerator from \cite{xu2023airvo} uses FP16 precision.

In general, ICE-BA integrated with FAST GPU accelerators is more high-performance, energy efficient, and accurate than ICE-BA with Harris GPU accelerators. However, ICE-BA+Vitis FAST shows slightly worse performance and energy efficiency than ICE-BA+Vitis Harris, also being less accurate in "easy" and "medium" sequences such as MH01, MH02, and MH03. Furthermore, despite being the configuration that yields the worst run-time performance and energy efficiency, ICE-BA+SuperPoint is not always more accurate than either ICE-BA+FAST or ICE-BA+Harris. For example, on both hardware platforms, ICE-BA+SuperPoint is only more accurate than ICE-BA+FAST and ICE-BA+Harris on the MH01 "easy" sequence that has good texture and illumination.

We also discovered that the use of hardware accelerators for feature detection, might have a positive effect on run-time performance for the downstream ICE-BA modules in its mapping thread, especially for the global bundle adjustment module. Table \ref{table:iceba_frontend_backend_fast} summarizes the run-time of the feature detection module and the localization thread, as well as the run-time and accuracy of the ICE-BA pipeline integrated with different FAST implementations. According to Table \ref{table:iceba_frontend_backend_fast}, after replacing the FAST implementation from OpenCV with FTFast and Vitis FAST, the decrease in run-time for the feature detection module is 1.01 ms - 2.15 ms (GPU) and 4.72 ms - 7.12 ms (FPGA), respectively, while the decrease in run-time for the localization thread is 0.61 ms - 1.19 ms (GPU) and 3.4 ms - 4.93 ms (FPGA), respectively. However, the decrease in run-time for the pipeline is much larger, i.e., 15.29 ms - 63.34 ms (GPU) and 63.91 ms - 274.46 ms (FPGA), respectively. Considering the localization thread runs in parallel with the local and global bundle adjustment modules in the mapping thread, and the global bundle adjustment module is empirically the most time-consuming module within a V-SLAM pipeline \cite{liu_ice-ba_2018}, we believe that the use of hardware accelerators for feature detection can affect the performance of the global bundle adjustment module. Further investigation shows that, when using FTFast or Vitis FAST, global bundle adjustment is invoked less frequently, compared with using OpenCV FAST, which leads to a decrease in run-time. Global bundle adjustment is a non-linear least squares system solver that jointly optimizes all the landmarks and the feature points that can be observed from each landmark in the global map, to further reduce the accumulated translation and rotation error, thus improving accuracy. It is surprising to find that, despite invoking global bundle adjustment less frequently, ICE-BA+FTFast is more accurate than ICE-BA+OpenCV FAST across all MH sequences. Although ICE-BA+Vitis FAST is, in general, less accurate than the software baseline, we believe this is because of the use of reduced-precision numbers and approximations in the Vitis FAST accelerator design.

\section{Conclusions}
\label{sec:conclusion}
This paper is the first study of feature detectors considering a V-SLAM on state-of-the-art SoCs with FPGA/GPU. The evaluation shows that when using a non-learning-based feature detector such as FAST and Harris, FTFast and Harris from the Nvidia VPI library, as well as ICE-BA+FTFast and ICE-BA+VPI Harris, can achieve better run-time performance and energy efficiency than the Vitis FAST and Harris accelerators as well as the FPGA-accelerated ICE-BA. However, when considering a learning-based detector such as SuperPoint, the Vitis SuperPoint accelerator can achieve better run-time performance and energy efficiency (up to 3.1$\times$ and 1.4$\times$ improvements, respectively) than its GPU counterpart. ICE-BA+Vitis SuperPoint can also achieve comparable run-time performance compared to ICE-BA integrated with the SuperPoint GPU accelerator, with better FPS in 2 out of 5 dataset sequences. However, when considering the accuracy, the results show that the GPU-accelerated ICE-BA is more accurate than the FPGA-accelerated ICE-BA in general. We also discovered that the use of hardware acceleration for feature detection could further improve the run-time of the V-SLAM pipeline by having global bundle adjustment (typically the most time-consuming module) invoked less frequently, while not sacrificing the accuracy.


%%%%%%%%%%%%%%%%%%%%%%%%%%%%%%%%%%%%%%%%%%%%%%%%%%%%%%%%%%%%%%%%%%%%%%%%%%%%%%%%

%%%%%%%%%%%%%%%%%%%%%%%%%%%%%%%%%%%%%%%%%%%%%%%%%%%%%%%%%%%%%%%%%%%%%%%%%%%%%%%%

%%%%%%%%%%%%%%%%%%%%%%%%%%%%%%%%%%%%%%%%%%%%%%%%%%%%%%%%%%%%%%%%%%%%%%%%%%%%%%%%
%\section*{APPENDIX}


\section*{Acknowledgement}

This work is partially funded by the UK Industrial Strategy Challenge Fund (ISCF) under the Digital Security by Design (DSbD) Programme delivered by UKRI as part of the Soteria (75243) projects and EPSRC EP/T026995/1 (EnnCore project). Mikel Luj\'an is supported by a Royal Society Wolfson Fellowship and an Arm/RAEng Research Chair Award.

%%%%%%%%%%%%%%%%%%%%%%%%%%%%%%%%%%%%%%%%%%%%%%%%%%%%%%%%%%%%%%%%%%%%%%%%%%%%%%%%

\printbibliography

\end{document}